\documentclass{article}
\usepackage{graphicx} 
\usepackage{textgreek}
\usepackage{array}
\usepackage{listings}
\graphicspath{ {C:/code/paperimages} }

\title{Comparative Analysis of CHATGPT and the evolution of language models}
\author{$^1$$^,$$^2$Oluwatosin Ogundare, $^1$Gustavo Quiros Araya}
\date{%
    $^1$Siemens Technology\\%
    $^2$California State University, San Bernardino\\[2ex]%
    \today
}

\begin{document}

\maketitle

\begin{abstract}
   Interest in Large Language Models (LLMs) has increased drastically since the emergence of ChatGPT and the outstanding positive societal response to the ease with which it performs tasks in Natural Language Processing (NLP). The triumph of ChatGPT, however, is how it seamlessly bridges the divide between language generation and knowledge models. In some cases, it provides anecdotal evidence of a framework for replicating human intuition over a knowledge domain. This paper highlights the prevailing ideas in NLP, including machine translation, machine summarization, question-answering, and language generation, and compares the performance of ChatGPT with the major algorithms in each of these categories using the Spontaneous Quality (SQ) score. A strategy for validating the arguments and results of ChatGPT is presented summarily as an example of safe, large-scale adoption of LLMs.
\end{abstract}
\section{Introduction to Classical Theories in Computational Linguistics}
A modern history of language models must admit that computational reasoning over any grammar or a set of symbols necessitates structure. Ferdinand de saussure's doctrine on structuralism lays out the key concepts of a scientific model of language with the view that it encapsulates a finite set of symbols that is independent of the intention or the psychological process that generates them \cite{radford2005structuralism}. Noam Chonksky improved on this idea by positing that all humans share an innate cognitive grammar structure and that this universal grammar structure generalizes principles on which language derives it's form. This idea forms the basis of generative grammar. Consequently, language can be modelled as a state machine that has infinite realization from a generation and deciphering standpoint \cite{chomsky2002syntactic}. A main result of generative grammar in terms of the theory of computational linguistics is the introduction of a formal framework on which logical operations can be performed on linguistics objects known as the minimalist framework \cite{chomsky2014minimalist}. P.W. Culicover and L.McNaly advocates for more sophistication in the tools used for analysis in generative syntax and potentially other ideas that have emerged from Chomsky's generative grammar, an indication that there is a limitation in what can be acheived by models that rely on the minimalist framework for formal linguistic reasoning \cite{culicover1998exploring}. For example, classical linguistic ideas like anaphora, an expression whose meaning depends on the antecedent or the context established by the antecedent, remains difficult within the framework of the minimalist program. Other theories that investigate the semantic meaning of language like functional linguistics improved the state of the art by introducing its own formalism that encompasses taxonomy and the notation that analyzes the structure of language which tend to perform better than the minimalist program when `meaning' or `intent' is the principal focus of a language task \cite{halliday2013halliday}. Another major early contribution to computational language analysis is Part-of-Speech (POS) tagging. At its inception, many of the approaches were rule based but famously in 1989, Kenneth Church automated POS tagging using a Hidden Markov Model (HMM) \cite{church1989stochastic}. Part-of-Speech tagging remains a major Natural Language Processing (NLP) task in both syntax and semantic analysis of text. A direct consequence of automatic POS tagging is the idea of classifying the nouns based on the domain of reference or generally based on language specific roles. This type of NLP task is referred to as Named Entity Recognition (NER). Named Entity Recognition identifies the subjects and objects of sentences based on the position of certain parts of speech as shown by Bikel, Miller \& Schwartz in 1998 \cite{bikel1998nymble}. Other techniques for NLP tasks quickly developed with the success of stochastic approaches to language modelling. For example, Ngrams was used widely in the early days for text categorization and still remains popular in modern NLP models\cite{suen1979n}. More recently stochastic approaches have very concrete applications in evaluating document similarity, performing semantic role labelling and other mainstream NLP tasks \cite{compositedocs}.

\section{ChatGPT and Prevailing Ideas in NLP}
The landscape of prevailing computational theories in Natural Language Processing (NLP) shifts in response to what constitute the popular NLP tasks. Currently, the major tasks are Machine Translation, Machine Summarization, Question Answering and Language Generation. There are minor tasks like text classification or control program generation that can be solved efficiently using simple statistical learning or machine learning models when the domain is well understood\cite{fbdprogram}. Although the quality of the performance of mainstream algorithms over the set of identified tasks have progressively improved through the decades, especially since the advent of deep learning, none has been able to perform quite as well over the all of the major tasks like ChatGPT. To demonstrate this claim, we select the major mainstream algorithm in every category and compare the performance of the chosen algorithm against ChatGPT using what we refer to as the Spontaneous Quality (SQ) score which is a measure of perceived quality, i.e., how well the output reflects the intention at first glance without in-depth linguistic analysis. The Spontaneous Quality score computes a value for a reference language model by accessing its performance over 3 criteria.
\begin{enumerate}
  \item Accuracy
  \item Clarity
  \item Native Speaker Likeness
\end{enumerate}
For scale in-variance, the accuracy score, the clarity score and Native Speaker Likeness score are on the interval (0, 1) and weighted in the SQ score by \{$\alpha_{i}$\}.\\ \\
Mathematically,\\ \\
 . \hspace{1cm} SQ Score = $\sum_{i} \alpha_{i} P_{i}$\\ \\
 .\hspace{3cm} Where $\alpha_{i} =$ Coefficient of Importance\\
 . \hspace{4cm} $P_{i} =$ Performance criteria\\ \\
 The reference test selects a paradigmatic average case in a mainstream language and another archetype of the average case in an obscure language. The main idea is to construct a shallow normalized performance metric, that can be repeated and averaged over a series of random pairs of mainstream and obscure languages. The reference test score is computed as the average of the series of geometric mean of the Spontaneous Quality (SQ) score of the algorithm in the mainstream case and the Spontaneous Quality (SQ) score of the algorithm in the obscure case.\\
 Mathematically,\\ \\
 Reference Test (RT) Score = $\frac{1}{k}\sum_{i=1}^{k}\sqrt{(SQ \hspace{1mm}Score_{main}^{i} \times SQ\hspace{1mm}Score_{obscure}^{i})}$

 \subsection{Machine Translation}

 REFERENCE TEST\\ 
 Using coefficients of importance { 0.5, 0.25, 0.25} over the performance criteria {accuracy, clarity, native speaker likeness}
 \begin{itemize}
     \item Mainstream case $:: English \rightarrow Spanish$
     \item Obscure case    $:: English \rightarrow Yoruba$
 \end{itemize}
 Test Phrase: "The man is a man that is a unique man"
 \\ 
 Spanish:
  \begin{itemize}
     \item Google Translate:: "El hombre es un hombre que es un hombre único."
     \item Chat GPT:: "El hombre es un hombre que es un hombre único."
 \end{itemize}
 Both translations are identical with the same SQ score.\\
 SQ Score = $0.5 \times 1 + 0.25 \times 1 + 0.25 \times 1 = 1$\\
 Yoruba:
   \begin{itemize}
     \item Google Translate:: "Okunrin naa je okunrin ti o je eniyan alailegbe"
     \item Chat GPT:: "Okunrin ni okunrin ti o ni okunrin kan pupo"
 \end{itemize} 
 SQ Score for Google Translate = $0.5 \times 0.9 + 0.25 \times 0.8 + 0.25 \times 0.5 = 0.725$\\
SQ Score for Chat GPT = $0.5 \times 0.3 + 0.25 \times 0.2 + 0.25 \times 0.7 = 0.375$\\
For a series of 1 language pair, the RT score is simply the geometric mean as shown below:\\
The Reference Test (RT) Score for Google Translate = $\sqrt{1 * 0.725}$ = 0.8514 \\
The Reference Test (RT) Score for Chat GPT = $\sqrt{1 * 0.375}$ = 0.6123 \\ \\
SUMMARY OF GOOGLE TRANSLATE VS CHATGPT OVER SELECT LANGUAGES
\begin{table}[h!]
\begin{center}
\begin{tabular}{| m{2cm} | m{2cm}| m{1cm} | m{2cm} | m{1cm}| m{1.5cm} |} 
  \hline
  Model & Language & SQ Score & Language & SQ Score & Instance RT Score \\
  \hline
  Google Translate & Russian & 0.95 & Tajik & 0.9 & 0.925 \\
  \hline
  Chat GPT & Russian & 1.0 & Tajik & 0.85 & 0.922 \\
  \hline
  Google Translate & French & 1.0 & Hungarian & 0.9 & 0.949 \\
  \hline
  Chat GPT & French & 1.0 & Hungarian & 1.0 & 1.0\\
    \hline
  Google Translate & Hindi & 1.0 & Telugu & 0.7 & 0.837 \\
  \hline
  Chat GPT & Hindi & 0.9 & Telugu & 0.7 & 0.794 \\
    \hline
  Google Translate & Arabic & 0.75 & Hausa & 0.7 & 0.725 \\
  \hline
  Chat GPT & Arabic & 0.75 & Hausa & 0.1 & 0.274 \\
   \hline
\end{tabular}

\caption{Performance of Google Translate VS ChatGPT in Machine Translation.}
\label{table:1}
\end{center}
\end{table}
\\The RT score for Google Translate over 4 language pairs = 0.969\\
The RT Score for ChatGPT over 4 language pairs = 0.748
 \subsection{Machine Summarization}
  REFERENCE TEST\\
  The reference text is the last section of the well known poem titled "Piano" by D.H. Lawrence
  \begin{center}
  "So now it is vain for the singer to burst into clamour
With the great black piano appassionato. The glamour
Of childish days is upon me, my manhood is cast
Down in the flood of remembrance, I weep like a child for the past"
  \end{center}
    \begin{itemize}
     \item QuillBot.com Summarizer:: "The singer's manhood is cast aside and he weeps for the past."
     \item Chat GPT:: "As I recall the memories of my childhood, I am overcome with nostalgia and weep uncontrollably, feeling as though my manhood has been washed away by the flood of reminiscence."
 \end{itemize}
 SQ Score for QuillBot.com Summarizer = $0.5 \times 0.6 + 0.25 \times 0.8 + 0.25 \times 0.8 = 0.7$\\
SQ Score for Chat GPT = $0.5 \times 0.9 + 0.25 \times 1 + 0.25 \times 1 = 0.95$\\ \\
The reference text for the obscure case is taken from D.O. Fagunwa's "Ireke Onibudo", P.11
  \begin{center}
  "Sugbon ibiti a de naa dara tobe ti mo feran ma le pe e ni ile rara,
  oniruurun ododo oloorunrun didun ni mo ri nibi sugbon ona ti won se si aarin awon ododo naa goolu ni won fi te ile ibe" \cite{ayoola2019critical} \cite{fagunwa_1949}
  \end{center}
      \begin{itemize}
     \item  QuillBot.com Summarizer:: "Ibiti a de naa dara, oniruuru ododo oloorunrun didun ni mo ri nibe."
     \item Chat GPT:: "Mo ni ki nfi ife oorun iwo kan bi aye mi, mo ni ki nse ilu mi, sugbon awon igbagbo ni won ni, niwaju gbogbo eto ni lo di okan, lati fi ogun isoro gbogbo ranmo ni ile."
 \end{itemize}
SQ Score for QuillBot.com Summarizer = $0.5 \times 0.65 + 0.25 \times 0.9 + 0.25 \times 0.8 = 0.75$\\
SQ Score for Chat GPT = $0.5 \times 0.3 + 0.25 \times 0.2 + 0.25 \times 0.4 = 0.3$\\ \\
The RT Score for  QuillBot.com Summarizer = $\sqrt{0.7 * 0.75}$ = 0.724, 
The RT Score for Chat GPT = $\sqrt{0.95 * 0.3}$ = 0.534 \\ \\
SUMMARY OF QuillBot.com SUMMARIZER VS CHATGPT OVER SELECT LANGUAGES

\begin{table}[h!]
\begin{center}
\begin{tabular}{| m{2cm} | m{2cm}| m{1cm} | m{2cm} | m{1cm}| m{1.5cm} |} 
  \hline
  Model & Language & SQ Score & Language & SQ Score & Instance RT Score \\
  \hline
  QuillBot.com & Russian & 0.9 & Tajik & 0.35 & 0.561 \\
  \hline
  Chat GPT & Russian & 0.75 & Tajik & 0.65 & 0.698 \\
  \hline
  QuillBot.com & French & 0.75 & Hungarian & 0.8 & 0.774 \\
  \hline
  Chat GPT & French & 0.8 & Hungarian & 0.9 & 0.849 \\
    \hline
  QuillBot.com & Hindi & 0.2 & Telugu & 0.025 & 0.071 \\
  \hline
  Chat GPT & Hindi & 0.9 & Telugu & 0.9 & 0.9 \\
    \hline
  QuillBot.com & Arabic & 0.65 & Hausa & 0.8 & 0.721 \\
  \hline
  Chat GPT & Arabic & 0.9 & Hausa & 0.025 & 0.15 \\
   \hline
\end{tabular}

\caption{Performance of QuillBot.com Summarizer VS ChatGPT in Machine Translation.}
\label{table:2}
\end{center}
\end{table}
The RT score for QuillBot.com Summarizer and ChatGPT over 4 language pairs respectively is 0.531, 0.649
 \subsection{Language Generation}
 One of the triumphs of ChatGPT is how beautifully it produces words that are not unusual within the vocabulary of a target language and for mainstream languages as shown prior, it produces sentences that follow the rules of grammar in those languages.
Even in some obscure cases when it generates sentences whose meaning are divergent from the intent, it follows the rules of grammar and it strives to use words that are not unusual in the local vernacular. In contrast to natural language, ChatGPT performs consistently well when generating programs across a variety of programming languages and paradigms. In this section, we will select a reference test that generates a program in a mainstream programming language and see if ChatGPT's performance degrade when it generates PLC programs for industrial automation.\\ \\
 REFERENCE TEST\\
 Generate a control program for a fast acting valve in a chemical plant in python as well as Siemens S7 ladder logic.\\
 \\ \textbf{Chat GPT Python:}
 \begin{lstlisting}
# Fast-Acting Valve Control Program
import time
# Set the initial state of the valve to closed
valve_open = False
# Define the function to control the valve
def control_valve(open_time):
    global valve_open
    valve_open = True
    # Send command to open the valve
    print("Opening valve...")
    # Wait for the open_time duration
    time.sleep(open_time)
    valve_open = False
    # Send command to close the valve
    print("Closing valve...")
\end{lstlisting}
\vspace{2.5mm}
 \textbf{Chat GPT S7 Ladder Logic:}
\begin{lstlisting}
Program: Fast-Acting Valve Control Program

Step 1: Inputs
I0.0: Start button
I0.1: Stop button

Step 2: Outputs
Q0.0: Valve

Step 3: Program

LD R0, I0.0         ; Load input I0.0 into register R0
LD R1, I0.1         ; Load input I0.1 into register R1
LD Q0.0, 0          ; Set the valve to closed
TND 1.0s            ; Debouncing timer for the start button
LD R2, I0.0         ; Load input I0.0 into register R2
AND R3, R2, NOT R0  ; Detect rising edge of start button press
LD R4, I0.1         ; Load input I0.1 into register R4
AND R5, R4, NOT R1  ; Detect rising edge of stop button press
XOR R6, R3, R5      ; Check if either start or stop button has been pressed
ANB R7, R6, Q0.0    ; Check if valve is currently closed
ORM Q0.0, R7        ; Open valve if it is closed
TND 1.0s            ; Debouncing timer for the stop button
LD R8, I0.1         ; Load input I0.1 into register R8
AND R9, R8, NOT R4  ; Detect rising edge of stop button press
ANB Q0.0, NOT R9    ; Close valve if stop button is pressed
END
\end{lstlisting} 

We compute the SQ score for ChatGPT for the generated algorithms factoring the lack of detail in the problem statement as follows:\\ \\
SQ Score for ChatGPT Python = $0.5 \times 1 + 0.25 \times 0.9 + 0.25 \times 0.9 = 0.95$\\
SQ Score for Chat GPT = $0.5 \times 1 + 0.25 \times 1 + 0.25 \times 1 = 1$\\
The Reference Test (RT) Score over the 2 cases = 0.975\\ \\
SUMMARY OF CHATGPT PERFORMANCE IN GENERATING SOURCE OVER SELECT PROGRAMMING LANGUAGES
\begin{table}[h!]
\begin{center}
\begin{tabular}{| m{2cm} | m{2cm}| m{1cm} | m{2cm} | m{1cm}| m{1.5cm} |} 
  \hline
  Model & Language & SQ Score & Language & SQ Score & Instance RT Score \\
  \hline
  Chat GPT & C & 0.95 & Fortran77 & 0.95 & 0.95 \\
  \hline
  Chat GPT & CSharp & 0.95 & Rust & 0.95 & 0.95 \\
  \hline
  Chat GPT & Javascript & 0.95 & Ruby & 0.95 & 0.95 \\
  \hline
  Chat GPT & Haskell & 1.0 & FSharp & 1.0 & 1.0 \\
    \hline
  Chat GPT & Scala & 1.0 & Lisp & 1.0 & 1.0 \\
  \hline
  Chat GPT & Java & 0.95 & Golang & 0.95 & 0.95 \\
    \hline

   \hline
\end{tabular}
\caption{Performance of ChatGPT in Language Generation.}
\label{table:3}
\end{center}
\end{table} \\
The RT score for ChatGPT over 6 language pairs respectively is 0.967
\section{Information differentiation Strategy in Knowledge Models}
We introduce the notion of the mediator within the machine generation workflow, as shown in fugure 1, as a crucial step in ensuring the quality and ethical standards of the generated content. The mediator acts as a gatekeeper to check the output of the LLM chatbot against various guidelines, regulations, policy, design recommendations, and prevailing sanctions. These guidelines may include legal and ethical frameworks such as data protection laws, hate speech regulations, or product-specific guidelines. By verifying the generated content against these guidelines, the mediator ensures that the content is compliant with the relevant regulations.

In addition to checking against guidelines, the mediator also checks the output against the constraints imposed by other engineering or product artifacts. These constraints may include technical limitations or product-specific requirements. For example, a chatbot designed for a specific platform may need to follow certain design standards or incorporate specific features. By ensuring that the output is consistent with these constraints, the mediator ensures that the generated content is technically feasible and aligns with the product's goals.

If the generated content fails to pass any of these mandatory checks, the mediator raises a review flag, and the output is automatically rejected. This automatic rejection ensures that any content that does not meet the established standards is caught early in the process, minimizing the potential negative impact. If the output passes all the checks, the mediator approves the LLM output for further downstream processing, including integration with other systems, deployment, or publication. 

\begin{figure}[ht!]
    \centering
    \includegraphics[width=\textwidth]{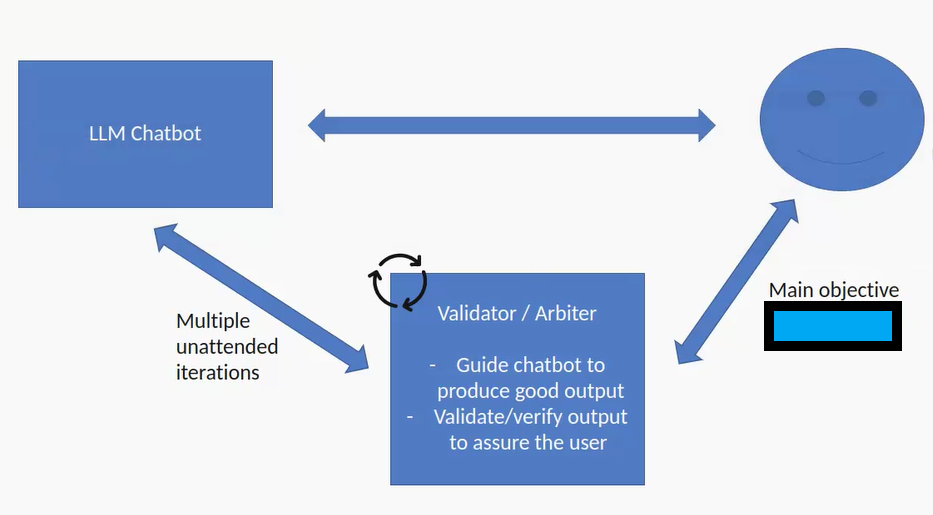}
    \caption{A process mediator working in tandem with an LLM.}
    \label{fig:overview}
\end{figure}

\section{Conclusion \& Further Work}
In conclusion, language models have come a long way since the early days of computational linguistics. From the introduction of generative grammar to the current state-of-the-art Large Language Models (LLMs) like ChatGPT, the field has made a lot of improvements in understanding and modeling language as well as bridging the gap between language models and knowledge models. With an increasing amount of available data for systematic training and the advancements in the computational theory of machine learning, we can expect language models to continue to improve and push the boundaries of what is possible in natural language processing especially applications in industrial automation. Future research on Large Language Models (LLMs) should investigate ethical concerns regarding the expanded functionality of LLMs as well attempt an explanation of the performance improvements in language generation. Additionally, a computational formalism of the notion of spontaneous human-like intuition that is sometimes evident in how ChatGPT performs on machine summarization would aid widespread adoption.

\bibliographystyle{plain} 
\bibliography{refs} 
\end{document}